\documentclass{article}

\usepackage{arxiv}

\usepackage[utf8]{inputenc} 
\usepackage[T1]{fontenc}    
\usepackage{hyperref}       
\usepackage{url}            
\usepackage{booktabs}       
\usepackage{amsfonts}       
\usepackage{nicefrac}       
\usepackage{microtype}      
\usepackage{lipsum}
\usepackage{graphicx}

\usepackage{tabularx}
\usepackage{dirtree}
\usepackage[most]{tcolorbox}
\usepackage{makecell}
\usepackage[switch]{lineno}
\usepackage[dvipsnames]{xcolor}
\usepackage{multirow}
\usepackage{setspace}
\usepackage{orcidlink}
\usepackage{placeins}

\newcommand{\yes}{\textcolor{ForestGreen}{\checkmark}}
\newcommand{\no}{\textcolor{red}{$\times$}}

\renewcommand{\shorttitle}{Synthetic 3D Gear Dataset (MFGNet-Gear)}

\title{A Synthetic 3D Gear Dataset for Manufacturing Quality Inspection (MFGNet-Gear)}

\author{
 Ruo-Syuan Mei {\normalfont\orcidlink{0009-0006-2259-952X}} \\
  Department of Mechanical Engineering\\
  University of Michigan\\
  Ann Arbor, MI 48109 \\
  \texttt{rsmei@umich.edu} \\
   \And
 Chenhui Shao {\normalfont\orcidlink{0000-0002-3299-2222}} \\
  Department of Mechanical Engineering\\
  University of Michigan\\
  Ann Arbor, MI 48109 \\
  \texttt{chshao@umich.edu} \\
}

\begin{document}
\date{July 10, 2026}
\setstretch{1.03}
\maketitle
\begin{abstract}
Quality control in smart manufacturing increasingly relies on data-driven methods, particularly deep learning, to automate the inspection of manufactured parts. Recent advances in three-dimensional (3D) metrology have enabled fine-scale assessment of dimensional accuracy, surface quality, and shape conformity. However, deep learning methods for point-cloud-based inspection require large volumes of labeled data covering part designs and defect types, which are costly and time-consuming to obtain. Moreover, defective parts are intrinsically rare in mass production, and the resulting class imbalance can degrade model performance and make rare defect types difficult to detect. Synthetic data generation (SDG) offers a promising approach to address these challenges by producing large, balanced, and fully annotated datasets. Yet, applying SDG to precision components requires representing part geometry and defect morphology parametrically, so that design and quality can be co-varied. This article describes MFGNet-Gear, a publicly available synthetic 3D dataset comprising 24,000 paired polygon meshes and point clouds across 12 gear designs and 4 quality classes, with 500 instances per design-quality combination. Gear geometries are generated with parametric computer-aided design software, with dimensional parameters perturbed by $\pm$0.0254 mm and defect parameters sampled from distributions representing defect morphologies. For each mesh, 100,000 points are uniformly sampled using Open3D and stored as N $\times$ 3 coordinate text files. Metadata labels identify the gear design and quality class, supporting part design classification, geometric defect detection, representation learning, and dataset benchmarking. MFGNet-Gear provides an open-source dataset for deep learning-based 3D metrology, with a reproducible generation pipeline extensible to additional part designs.
\end{abstract}

\vspace{0.5em}
\begin{center}
\begin{minipage}{0.85\textwidth}
\small
\textbf{Data DOI/PID:} \href{https://doi.org/10.7302/qrdj-n812}{doi.org/10.7302/qrdj-n812}\\
\\
\textbf{Data Type/Location:} 3D point clouds, polygon meshes; Deep Blue Data, University of Michigan
\end{minipage}
\end{center}
\vspace{0.5em}

\section{Background}

Geometric integrity directly impacts the functionality, reliability, and safety of final manufactured products, making part qualification based on geometric measurements a fundamental quality control activity in modern manufacturing \cite{pei2022review, buswell2022geometric}. Recent advancements in optical three-dimensional (3D) metrology have enabled high-resolution inspection of part geometry, generating 3D point cloud measurements that capture detailed surface geometry \cite{catalucci2022optical, huo2023research, yang2021data}. In precision components such as gears, small geometric deviations can reduce transmission efficiency and shorten service life \cite{fernandes1996tooth, fernandes1997surface}, making accurate detection of such deviations essential to quality assurance. To enable efficient 3D inspection at production scale, deep learning methods can support automated, end-to-end analysis of unstructured point cloud data.

Deep learning architectures for 3D point clouds, including PointNet \cite{qi2017pointnet} and PointNet++ \cite{qi2017pointnet++}, have driven advances in applications such as autonomous driving perception \cite{nekrasov2025spotting, yang2024visual} and robotic manipulation \cite{jia2025lift3d, fang2023anygrasp}. However, these methods may not directly transfer to manufacturing inspection, as manufacturing point clouds differ significantly from those in these domains in geometry, scale, and defect characteristics \cite{rani2024advancements}. Bridging this domain gap requires retraining models with manufacturing data. Yet such data are difficult to obtain because defective parts are scarce in mass production, datasets are highly imbalanced between conforming and defective samples, and annotation is costly and time-consuming \cite{cheng2026comprehensive, wang2023mvgcn, guo2025iec3d}. As a result, the use of deep learning for point-cloud-based manufacturing quality inspection remains underexplored.

Synthetic data generation (SDG) has emerged as a promising approach to address these data constraints by producing large-scale, balanced, and fully annotated datasets across multiple manufacturing inspection tasks and data modalities \cite{mei2025synthetic, buggineni2024enhancing, mei2026hybrid}. Realizing this promise for a given inspection task, however, depends on the availability of public benchmark datasets against which methods can be developed and compared. To our knowledge, no public 3D point-cloud dataset provides a gear benchmark that jointly varies part design and defect class (Table~\ref{tab:3dpc-datasets}). General benchmarks such as ModelNet40 \cite{wu2015modelnet}, ShapeNet \cite{chang2015shapenet}, and ScanObjectNN \cite{uy2019scanobjectnn} contain object categories rather than defect labels. Industrial anomaly detection datasets such as MVTec 3D-AD \cite{bergmann2021mvtec}, Real3D-AD \cite{liu2023real3d}, and Anomaly-ShapeNet \cite{li2024towards} capture surface anomalies but lack gear-specific design and defect labels. The two datasets closest in scope, Gear-PCNet++ \cite{xu2023defect} and IEC3D-AD \cite{guo2025iec3d}, target precision manufacturing components but are not publicly available, limiting reproducible benchmarking.

\begin{figure}[!htbp]
\centerline{\includegraphics[width=6.4in]{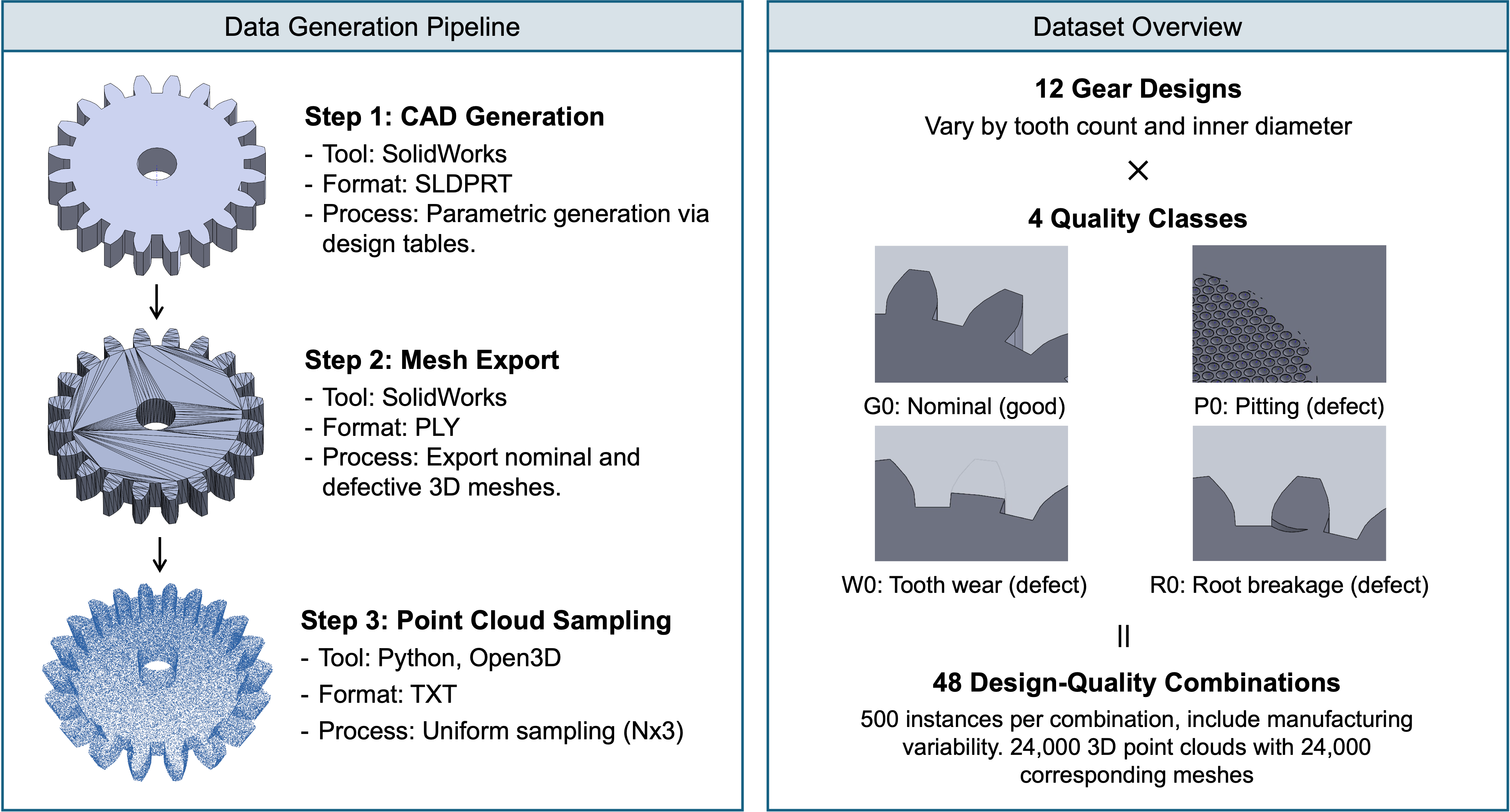}}
\caption{Overview of MFGNet-Gear dataset.}
\label{fig:MFGNet-Gear-Overview}
\end{figure}

To address this gap, this paper introduces MFGNet-Gear (Fig.~\ref{fig:MFGNet-Gear-Overview}), a publicly available synthetic 3D dataset comprising 24,000 paired point clouds and polygon meshes spanning 12 gear designs and 4 quality classes: good (G0), pitting (P0), tooth wear (W0), and tooth root breakage (R0), with 500 instances per design-class combination. Each point cloud contains 100,000 points sampled from the corresponding polygon mesh using Open3D, with geometric and defect parameters co-varied through design tables to introduce controlled part-to-part variation. An earlier study \cite{mei2024deep} used subsets of this dataset to train two models adapted from the PointNet++ architecture: a classifier for the 12 gear designs, and a defect detector for the 4 quality classes of a single design. That study reported design-classification accuracy up to 100\%, defect-detection accuracy of 85\% on a single design, and an optimal sampling resolution of 0.681 mm. MFGNet-Gear builds on this prior work by providing a reusable public benchmark, supporting the development and comparison of point-cloud deep learning methods across the full design–quality space.

\begin{table}[!htbp]
\caption{Comparison of representative 3D point-cloud datasets for manufacturing quality control. Mfg. indicates whether the dataset targets manufactured parts with production-relevant defect semantics.}
\label{tab:3dpc-datasets}
\small
\setlength{\tabcolsep}{4pt}
\renewcommand{\arraystretch}{1.2}

\begin{tabularx}{\textwidth}{
    >{\raggedright\arraybackslash}p{0.18\textwidth}
    >{\centering\arraybackslash}p{0.045\textwidth}
    >{\centering\arraybackslash}p{0.075\textwidth}
    >{\raggedright\arraybackslash}p{0.15\textwidth}
    >{\raggedright\arraybackslash}X
    >{\raggedright\arraybackslash}X
    >{\centering\arraybackslash}p{0.055\textwidth}
    >{\centering\arraybackslash}p{0.055\textwidth}
}
\toprule
\textbf{Dataset} &
\textbf{Year} &
\textbf{Type} &
\textbf{Data source} &
\textbf{Scale} &
\textbf{Defect types} &
\textbf{Mfg.} &
\textbf{Public} \\
\midrule

\multicolumn{8}{l}{\textit{General 3D benchmarks}} \\
\addlinespace[2pt]
ModelNet40 \cite{wu2015modelnet} 
    & 2015 & Synthetic & CAD; online repositories 
    & 12K models; 40 classes 
    & --- & \no & \yes \\

ShapeNet \cite{chang2015shapenet} 
    & 2015 & Synthetic & CAD; online repositories 
    & 51K models; 55 classes 
    & --- & \no & \yes \\

ScanObjectNN \cite{uy2019scanobjectnn} 
    & 2019 & Real & Depth sensor; indoor scans 
    & 2.9K objects; 15 classes 
    & --- & \no & \yes \\

\midrule
\multicolumn{8}{l}{\textit{Industrial anomaly detection datasets}} \\
\addlinespace[2pt]
MVTec 3D-AD \cite{bergmann2021mvtec} 
    & 2022 & Real & Structured-light scanner 
    & 4.1K scans; 10 classes 
    & Scratches, dents, holes 
    & \no & \yes \\

Anomaly-ShapeNet \cite{li2024towards} 
    & 2023 & Synthetic & ShapeNet + Blender 
    & 1.6K samples; 40 classes 
    & Generic deformations 
    & \no & \yes \\

Real3D-AD \cite{liu2023real3d} 
    & 2023 & Real & Structured-light scanner 
    & 1.3K items; 12 classes 
    & Surface anomalies 
    & \no & \yes \\

\midrule
\multicolumn{8}{l}{\textit{Gear and precision manufacturing datasets}} \\
\addlinespace[2pt]
Gear-PCNet++ \cite{xu2023defect} 
    & 2023 & Synthetic & CAD 
    & 10K samples; 5 designs 
    & Fracture, pitting, glue, wear 
    & \yes & \no \\

IEC3D-AD \cite{guo2025iec3d} 
    & 2025 & Real & Structured-light system 
    & 2.4K samples; 15 classes 
    & Protrusions, pits, scratches 
    & \yes & \no \\

\textbf{MFGNet-Gear (Ours)} 
    & \textbf{2026} 
    & \textbf{Synthetic} 
    & \textbf{CAD} 
    & \textbf{24K parts; 12 designs and 4 quality classes} 
    & \textbf{Pitting, tooth wear, tooth root breakage} 
    & \yes 
    & \yes \\

\bottomrule
\end{tabularx}
\end{table}

\FloatBarrier

\section{Collection Methods and Design}

\begin{figure}[!htbp]
\centerline{\includegraphics[width=6.5in]{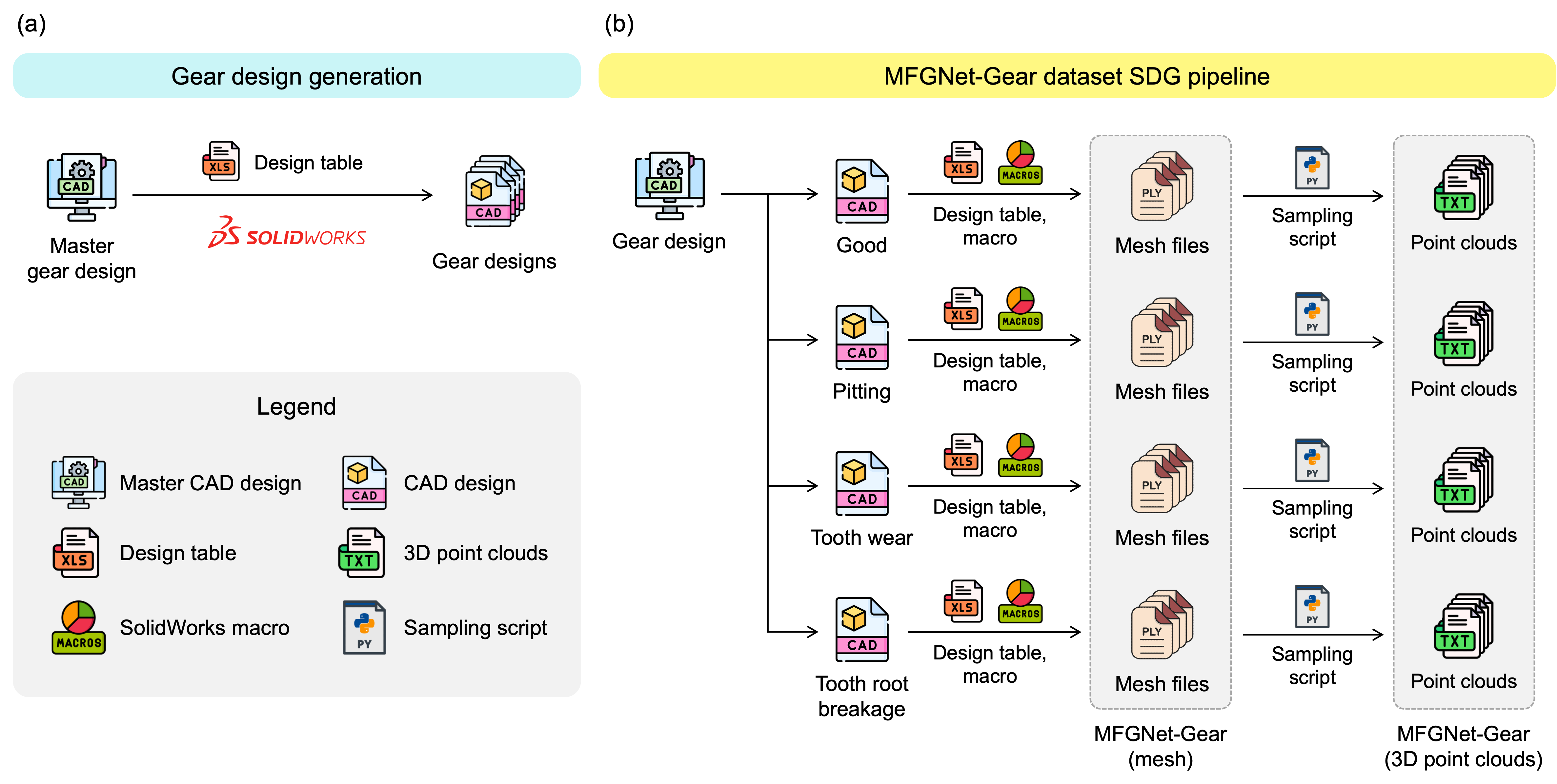}}
\caption{MFGNet-Gear SDG pipeline. (a) generation of 12 gear designs; (b) generation of mesh and 3D point clouds by jointly varying 4 quality classes.}
\label{fig:SDG}
\end{figure}

The MFGNet-Gear dataset is generated using a three-step pipeline: parametric computer-aided design (CAD) in SolidWorks 2021, mesh export in PLY format, and uniform surface sampling to generate 3D point clouds (Fig.~\ref{fig:SDG}). Each stage is described below to ensure reproducibility across all 12 gear designs and 4 quality classes.

\begin{figure}[!htbp]
\centerline{\includegraphics[width=3.6in]{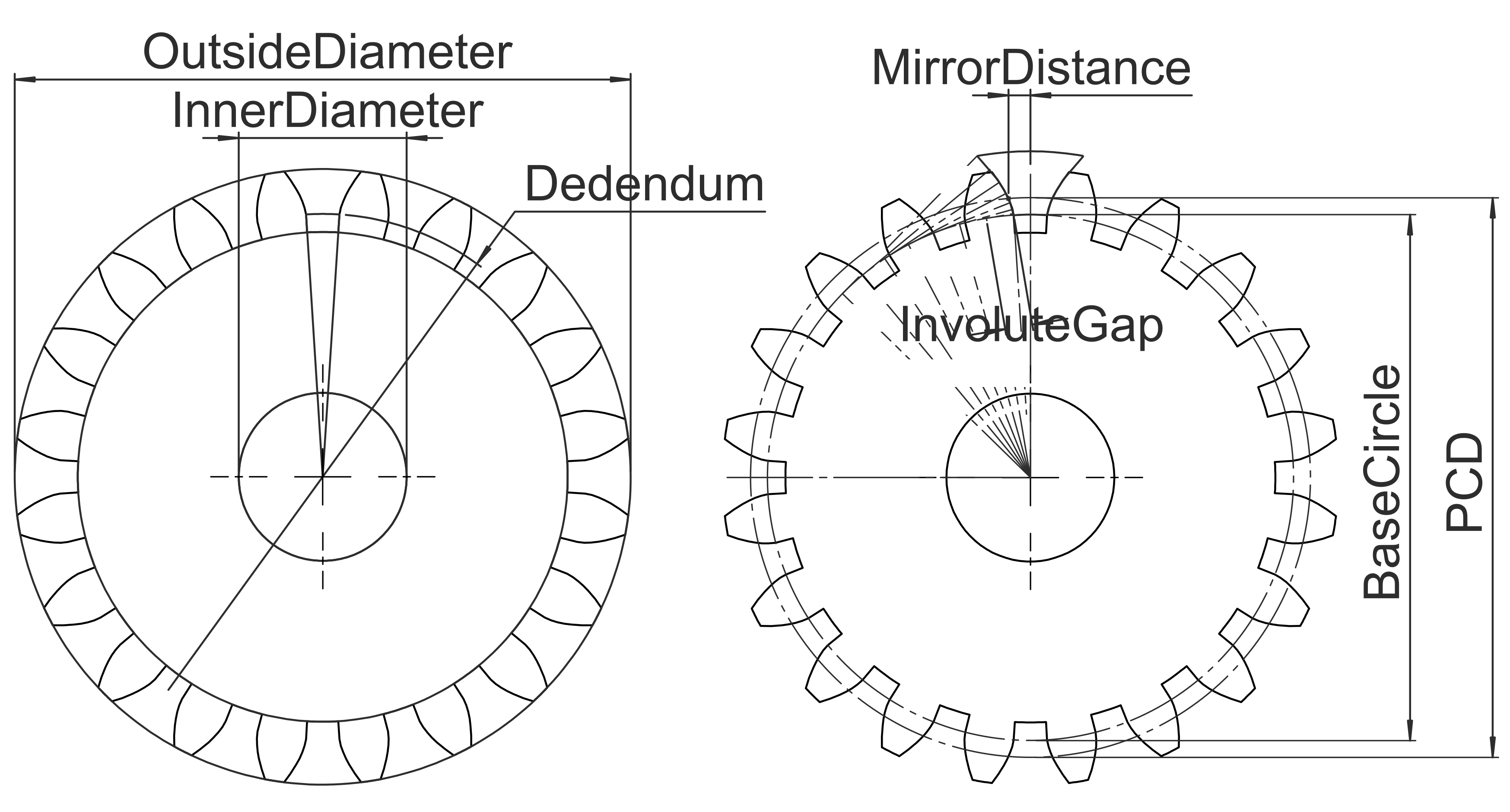}}
\caption{Schematic of a gear design.}
\label{fig:GearSchematic}
\end{figure}

\subsection{Gear design specification}
\begin{figure}[!htbp]
\centerline{\includegraphics[width=4.4in]{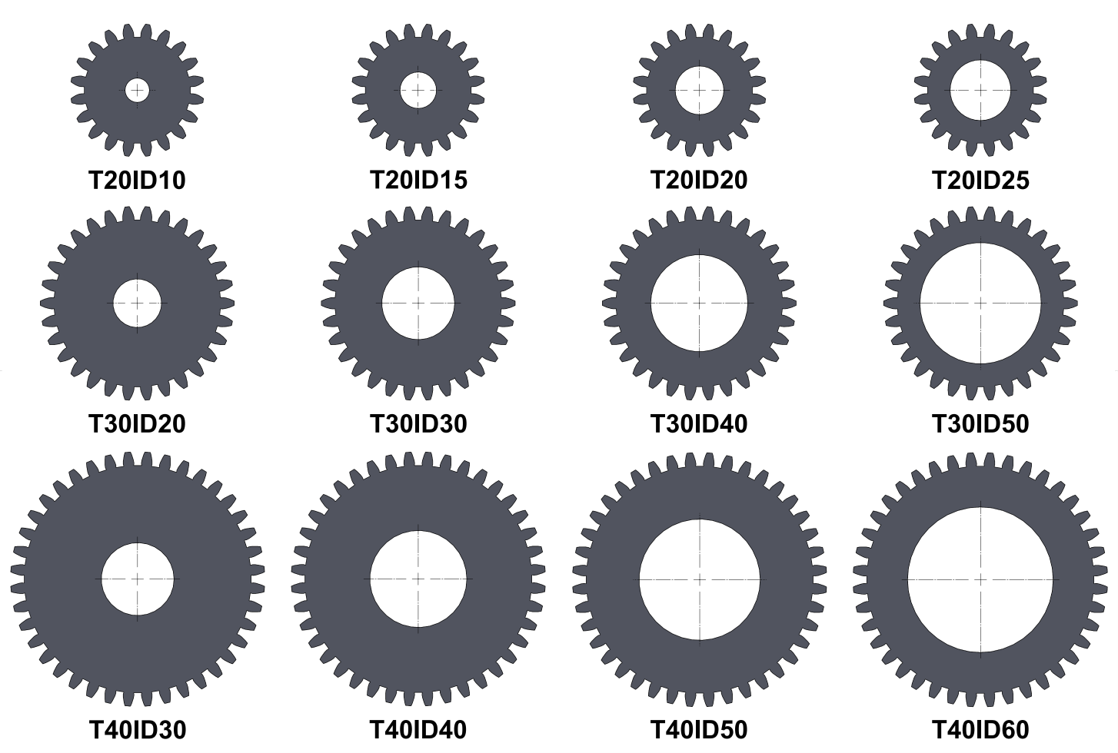}}
\caption{Overview of twelve gear designs.}
\label{fig-GearDesignOverview}
\end{figure}

Twelve gear designs are parameterized using eight geometric parameters: outside diameter (OD), base circle, pitch circle diameter, mirror distance, involute gap, dedendum (Den.), number of teeth, and inner diameter (ID), as shown in Fig.~\ref{fig:GearSchematic}. Values for each parameter are specified in Table~\ref{tab-12GearDesignTable}, yielding 12 unique configurations that span three tooth counts (20, 30, 40) and four inner diameters per tooth count, as illustrated in Fig.~\ref{fig-GearDesignOverview}. To represent dimensional variation from machining tolerances, a perturbation of $\pm$0.0254~mm (0.001~in) is applied to each geometric parameter in the design tables, with the magnitude informed by gear accuracy standards \cite{ISO1328-1-2013}. This perturbation introduces controlled part-to-part geometric variation rather than reproducing idealized CAD geometry.

\vspace{30pt}

\begin{table}[!htbp]
\caption{Design specification used to create twelve gear designs (mm).}
\label{tab-12GearDesignTable}
\centering
\setlength{\tabcolsep}{4.5pt}
\renewcommand{\arraystretch}{1.10}
\begin{tabular}{lcccccccc}
\toprule
\textbf{Design} &
\makecell{\textbf{Number}\\\textbf{of teeth}} &
\makecell{\textbf{Inner}\\\textbf{diameter}} &
\makecell{\textbf{Outer}\\\textbf{diameter}} &
\makecell{\textbf{Base}\\\textbf{circle}} &
\makecell{\textbf{Pitch circle}\\\textbf{diameter}} &
\makecell{\textbf{Mirror}\\\textbf{distance}} &
\makecell{\textbf{Involute}\\\textbf{gap}} &
\textbf{Dedendum} \\
\midrule
T20ID10 & 20 & 10 & 55  & 46.98 & 50  & 1.96 & 2.46 & 43.75 \\
T20ID15 & 20 & 15 & 55  & 46.98 & 50  & 1.96 & 2.46 & 43.75 \\
T20ID20 & 20 & 20 & 55  & 46.98 & 50  & 1.96 & 2.46 & 43.75 \\
T20ID25 & 20 & 25 & 55  & 46.98 & 50  & 1.96 & 2.46 & 43.75 \\
\addlinespace[2pt]
T30ID20 & 30 & 20 & 80  & 70.48 & 75  & 1.96 & 3.69 & 68.75 \\
T30ID30 & 30 & 30 & 80  & 70.48 & 75  & 1.96 & 3.69 & 68.75 \\
T30ID40 & 30 & 40 & 80  & 70.48 & 75  & 1.96 & 3.69 & 68.75 \\
T30ID50 & 30 & 50 & 80  & 70.48 & 75  & 1.96 & 3.69 & 68.75 \\
\addlinespace[2pt]
T40ID30 & 40 & 30 & 105 & 94.48 & 100 & 1.96 & 5.06 & 93.75 \\
T40ID40 & 40 & 40 & 105 & 94.48 & 100 & 1.96 & 5.06 & 93.75 \\
T40ID50 & 40 & 50 & 105 & 94.48 & 100 & 1.96 & 5.06 & 93.75 \\
T40ID60 & 40 & 60 & 105 & 94.48 & 100 & 1.96 & 5.06 & 93.75 \\
\bottomrule
\end{tabular}
\end{table}

\FloatBarrier

\subsection{Quality classes parameterization}

The dataset includes three common gear failure modes: pitting, tooth wear, and tooth root breakage. Pitting occurs when contact stress exceeds the surface fatigue limit, producing hemispherical cavities on the tooth surface \cite{fernandes1997surface}. Tooth wear results from friction-driven material removal along the contact path, progressively reducing tooth thickness \cite{wu1993sliding}. Tooth root breakage develops from stress concentration at the tooth root under overload, forming a wedge-shaped crack \cite{fernandes1996tooth}. Ten parameters define these three defect types, as specified in  Table~\ref{tab:3DefectiveGearDimension}.

Each parameter distribution reflects the physical characteristics of the corresponding failure mode. Pitting initiation is spatially random, so its location (pittingAngle, pittingDist) and size (pittingR, dentR) follow uniform distributions  $\mathcal{U}$. Tooth wear extent (ToothLossDist) is similarly sampled from $\mathcal{U}(0,\,0.5(OD-Den.))$, reflecting the variable progression of abrasive wear. Tooth root breakage parameters (ToothDepth, ToothDist1, ToothDist2) follow normal distributions $\mathcal{N}$. ToothWidth is fixed at 0.81 mm, as this dimension is constrained by the root geometry.

\begin{table}[!htbp]
\caption{Parameter settings used to generate three gear defect types (mm).}
\label{tab:3DefectiveGearDimension}
\centering
\setlength{\tabcolsep}{6pt}
\renewcommand{\arraystretch}{1.15}

\begin{tabular}{lll}
\toprule
\textbf{Defect type} & \textbf{Parameter} & \textbf{Value} \\
\midrule
\multirow{4}{*}{Pitting}
    & dentR       & $\mathcal{U}(0.05, 0.1)$ \\
    & pittingR    & $\mathcal{U}(1, 3)$ \\
    & pittingAngle & $\mathcal{U}(0, 360)$ \\
    & pittingDist & $\mathcal{U}\left(\frac{ID}{2}+\textit{pittingR}, \frac{Den.}{2}-\textit{pittingR}\right)$ \\
\midrule
Tooth wear
    & ToothLossDist & $\mathcal{U}\left(0, 0.5(OD-Den.)\right)$ \\
\midrule
\multirow{7}{*}{Tooth-root breakage}
    & ToothDepth & $\mathcal{N}(2.9, 0.3)$ \\
    & ToothDist1 & $\mathcal{N}(1.1, 0.2)$ \\
    & ToothDist2 & $\mathcal{N}(1.1, 0.2)$ \\
    & ToothWidth & $0.81$ \\
    & ToothTail & $\mathcal{U}(21.875, 0.117)$ for T20 gears \\
    & ToothTail & $\mathcal{U}(34.375, 0.117)$ for T30 gears \\
    & ToothTail & $\mathcal{U}(46.875, 0.117)$ for T40 gears \\
\bottomrule
\end{tabular}
\end{table}

\FloatBarrier

\subsection{3D mesh and point cloud generation}

Each gear CAD model is exported from SolidWorks 2021 as a polygon mesh in PLY format using a SolidWorks macro. Open3D is then used to uniformly sample 100,000 points from each mesh. The resulting point cloud is normalized to the unit sphere and saved as a comma-delimited TXT file containing N $\times$ 3 XYZ coordinates. The 100,000-point sampling density is comparable to the output of commercial industrial-grade 3D scanners and was shown in prior work \cite{mei2024deep} to capture geometric features sufficient for classification while remaining computationally tractable for PointNet-based architectures.

In total, the pipeline generates 48 design-quality combinations (12 designs $\times$ 4 classes) with 500 instances each, resulting in 24,000 meshes and 24,000 corresponding point clouds. The pipeline is implemented using Python scripts and SolidWorks macros. All scripts are provided in the dataset GitHub repository (\url{https://github.com/AliceRSMei/MFGNet-Gear}).

\section{Validation and Quality}

The quality of the MFGNet-Gear dataset is evaluated from four perspectives: completeness and class balance, file-level structural integrity, sampling fidelity, and geometric accuracy relative to the design specification. Because the dataset is synthetically generated rather than sensor-acquired, ``accuracy'' is quantified as the agreement between the generated geometry and the intended CAD design. All checks can be reproduced with the script released in the GitHub repository (Section \ref{sec:source-code}).

\begin{figure}[!htbp]
\centerline{\includegraphics[width=3in]{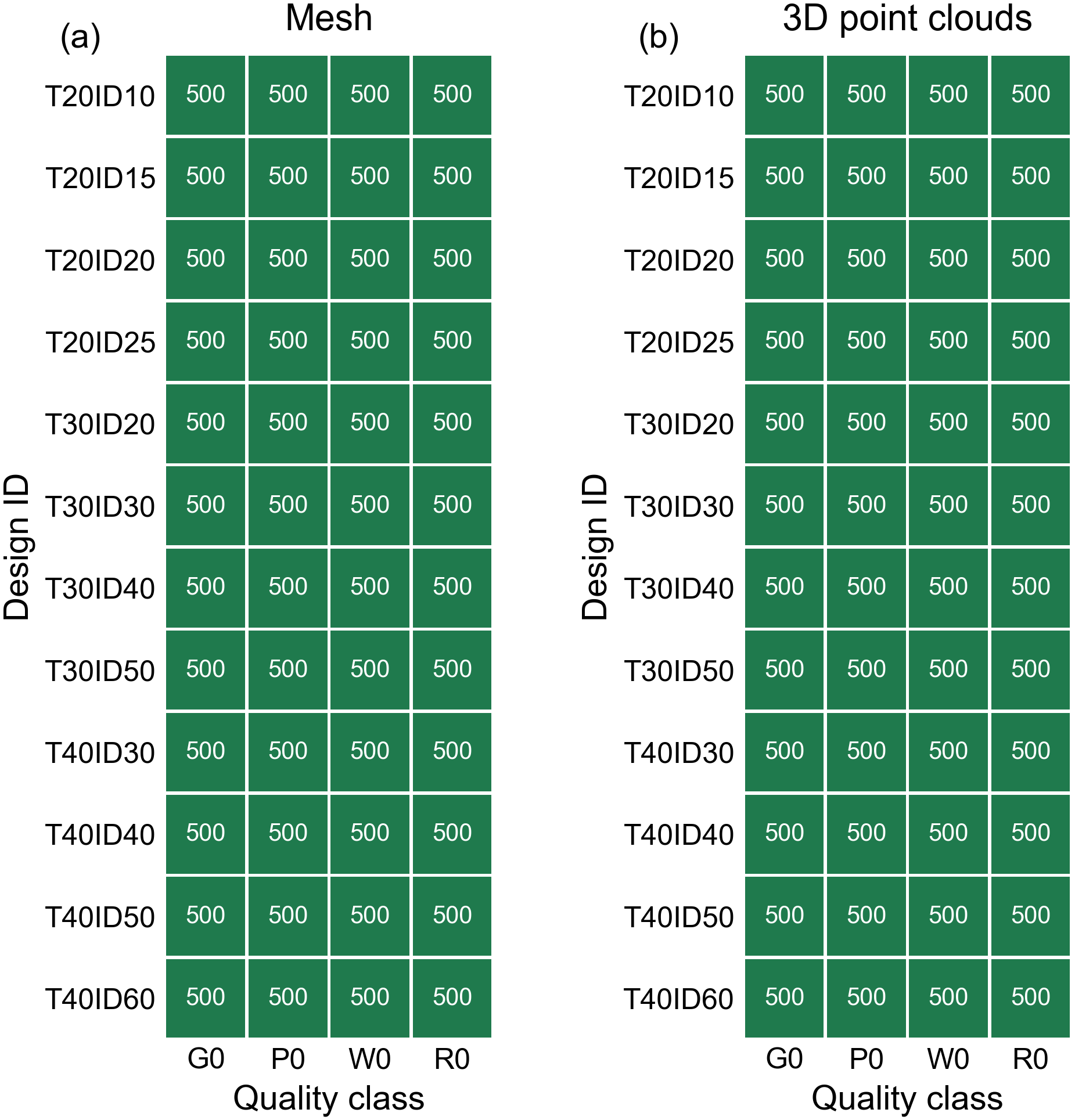}}
\caption{File counts per design-quality combination: (a) meshes; (b) 3D point clouds. All 48 combinations contain exactly 500 instances.}
\label{fig:dataset-file-count}
\end{figure}

To confirm that the dataset is complete and class-balanced, the validation script enumerated every file in the 48 design--class subfolders of \texttt{mesh\_ply} and \texttt{pointcloud\_txt} and checked, for each subfolder, both the instance count and the correspondence between mesh and point-cloud filenames. Each subfolder contains exactly 500 instances, and every mesh has an identically named point-cloud counterpart, with no missing or extra files (Fig.~\ref{fig:dataset-file-count}). The 12 gear designs $\times$ 4 quality classes are therefore complete and class-balanced, yielding 24,000 files in each format.

To confirm that every file is well-formed and machine-readable, each file was loaded and parsed: each point cloud was required to contain an $N\times3$ array of exactly 100,000 finite, unique, numeric coordinates, and each mesh was required to load as a valid PLY file representing a closed surface with no boundary edges or degenerate, zero-area faces. All 24,000 point clouds passed (100.0\%), and 23,990 of 24,000 meshes passed (99.96\%). The 10 meshes that did not pass are all in the pitting (P0) class: tessellating the sub-millimeter pit cavities, the finest geometric feature in the dataset, occasionally introduces a single near-zero-area triangle or one unclosed edge at the pit boundary. These artifacts do not affect point-cloud sampling, and all 24,000 meshes were sampled successfully. The 0.04\% shortfall reflects a benign meshing artifact at the smallest defect scale rather than an error in the underlying CAD geometry.

\begin{figure}[!htbp]
\centerline{\includegraphics[width=3.4in]{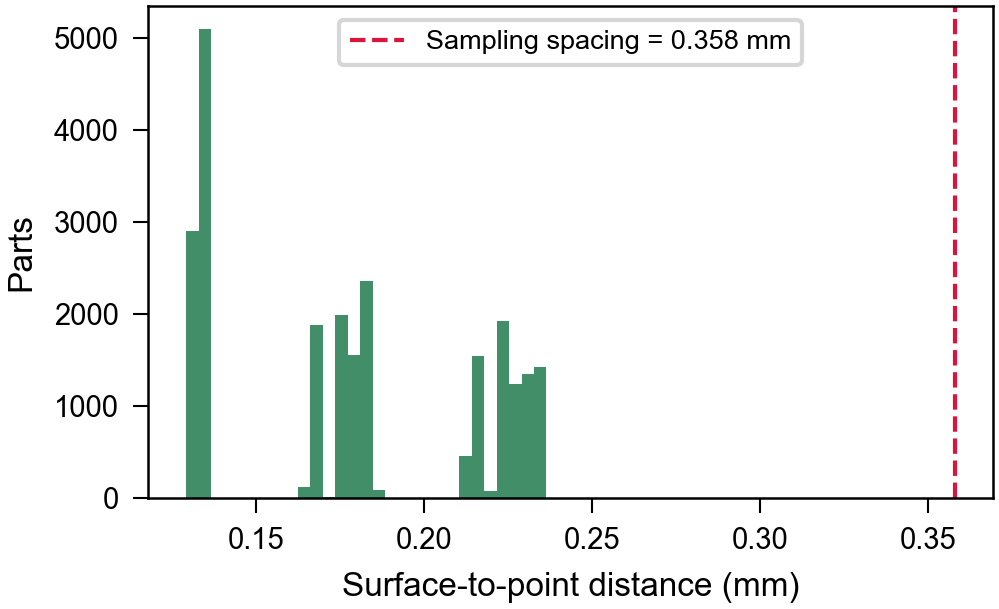}}
\caption{Mean surface-to-point distance over all 24,000 parts. All parts lie near half the theoretical sampling spacing (dashed), as expected for uniform coverage.}
\label{fig:point-to-surface}
\end{figure}

To confirm that the sampling faithfully and uniformly represents the CAD surface, the distance from the mesh surface to the nearest sampled point was measured for each mesh and compared with the theoretical inter-point spacing $\sqrt{A/N}$, where $N{=}100,000$ and $A$ is the surface area. Averaged over all 24,000 parts, the mean surface-to-point distance was 0.18~mm, approximately half the theoretical spacing of 0.36~mm expected for uniform sampling (Fig.~\ref{fig:point-to-surface}). This agreement confirms that the sampling covers the surface densely and evenly, without gaps or clustering.

\begin{figure}[!htbp]
    \centering
    \includegraphics[width=0.5\linewidth]{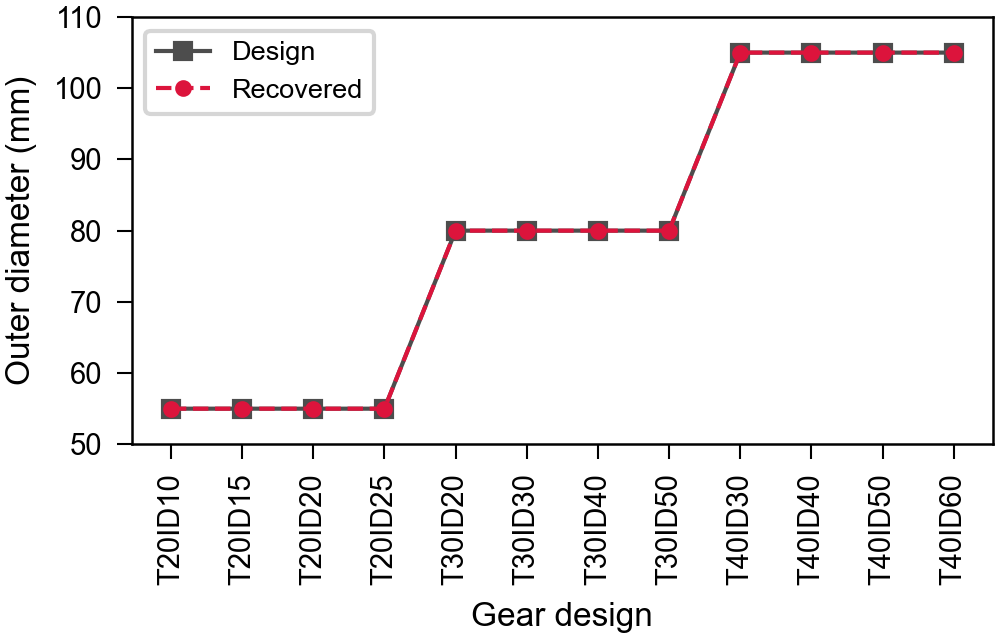}
    \caption{Recovered versus design OD for the twelve gear designs (MAE 0.007~mm). Tooth count is recovered exactly for all designs.}
    \label{fig:feature-recovery}
\end{figure}

Finally, to confirm that the generated geometry matched the intended design, two global features were recovered from each part and compared against the design table: the OD (computed as twice the maximum radius about the gear axis) and the tooth count (computed as the dominant angular period of the outer profile). Across all 12 designs, the OD was recovered with a mean absolute error (MAE) of 0.007~mm (maximum error of 0.035~mm), and the tooth count was recovered exactly (Table~\ref{tab:feature-recovery}, Fig.~\ref{fig:feature-recovery}). The OD remains recoverable in the defect classes because the represented defects remove material from the tooth flank or root rather than from the tooth tips. The tooth count is preserved in the released point clouds, as it is invariant to the applied unit-sphere normalization. The OD was verified on the millimeter-scale point clouds sampled from each mesh, as described in the validation script.

\begin{table}[!htbp]
\centering
\caption{Geometric accuracy for each design, evaluated using 2,000 instances (500 per quality class). Tooth count is recovered exactly (100\%); OD error is the mean absolute deviation from the nominal value.}
\label{tab:feature-recovery}
\begin{tabular}{lccrr}
\toprule
Design & Tooth count & Nominal OD & OD MAE & OD max error \\
       & (nominal = recovered) & (mm) & (mm) & (mm) \\
\midrule
T20ID10 & 20 & 55  & 0.0066 & 0.0269 \\
T20ID15 & 20 & 55  & 0.0067 & 0.0315 \\
T20ID20 & 20 & 55  & 0.0068 & 0.0283 \\
T20ID25 & 20 & 55  & 0.0067 & 0.0315 \\
T30ID20 & 30 & 80  & 0.0069 & 0.0321 \\
T30ID30 & 30 & 80  & 0.0067 & 0.0302 \\
T30ID40 & 30 & 80  & 0.0066 & 0.0295 \\
T30ID50 & 30 & 80  & 0.0067 & 0.0349 \\
T40ID30 & 40 & 105 & 0.0069 & 0.0324 \\
T40ID40 & 40 & 105 & 0.0068 & 0.0296 \\
T40ID50 & 40 & 105 & 0.0066 & 0.0305 \\
T40ID60 & 40 & 105 & 0.0068 & 0.0273 \\
\midrule
All     & --- & --- & 0.0067 & 0.0349 \\
\bottomrule
\end{tabular}
\end{table}

\section{Records and Storage}

The MFGNet-Gear dataset and the SDG pipeline are stored in two separate repositories. The dataset, comprising 3D gear meshes and corresponding 3D point clouds, is deposited in Deep Blue Data (\href{https://doi.org/10.7302/qrdj-n812}{doi.org/10.7302/qrdj-n812}). The SDG scripts are available in the GitHub repository described in Section \ref{sec:source-code}.

As shown in Fig.~\ref{fig:mfgnet-file-structure}, the MFGNet-Gear dataset is organized into a README file (\texttt{README.md}), a mesh folder (\texttt{/mesh\_ply}), and a point cloud folder (\texttt{/pointcloud\_txt}). The README file provides a dataset description and usage instructions. Mesh files are stored in PLY format in \texttt{/mesh\_ply}, and the corresponding point clouds are stored in \texttt{/pointcloud\_txt} as comma-delimited TXT files containing $N \times 3$ XYZ coordinates, normalized to the unit sphere.

Both folders share the same internal structure: 48 subfolders, each corresponding to one design-quality combination, and each containing 500 instances. For example, \texttt{/T20ID10G0} contains 500 nominal gears with 20 teeth and a 10~mm inner diameter, and \texttt{/T20ID10P0} contains 500 gears with pitting defects of the same design. Subfolder names follow the convention \texttt{T\{ToothCount\}ID\{InnerDiameter\}\allowbreak\{QualityClass\}0}, where \texttt{QualityClass} is one of \texttt{G} (good), \texttt{P} (pitting), \texttt{W} (tooth wear), or \texttt{R} (tooth root breakage). Within each subfolder, files follow the subfolder naming convention with a five-digit instance index (e.g., \texttt{T20ID10G0\_00001.txt}). Mesh and point cloud files for the same instance share identical filename stems across the two folders. Table~\ref{tab:dataset-summary} summarizes the dataset structure.

\begin{table}[!htbp]
\caption{Summary of the MFGNet-Gear dataset structure.}
\label{tab:dataset-summary}
\centering
\setlength{\tabcolsep}{8pt}
\renewcommand{\arraystretch}{1.15}

\begin{tabular}{lll}
\toprule
\textbf{Item} & \texttt{/mesh\_ply} & \texttt{/pointcloud\_txt} \\
\midrule
Format & PLY & TXT ($N \times 3$) \\
Subfolders & 48 & 48 \\
Files per subfolder & 500 & 500 \\
Total files & 24,000 & 24,000 \\
\bottomrule
\end{tabular}
\end{table}

\begin{figure}[!htbp]
\centering
\footnotesize
\begin{tcolorbox}[
    colback=gray!4,
    colframe=gray!25,
    boxrule=0.3pt,
    arc=1mm,
    left=2mm,
    right=2mm,
    top=1mm,
    bottom=1mm,
    width=0.88\linewidth
]
\dirtree{%
.1 \texttt{MFGNet-Gear/}.
.2 \texttt{README.md}.
.2 \texttt{mesh\_ply/}.
.3 \texttt{T20ID10G0/}.
.4 \texttt{T20ID10G0\_00001.ply}.
.4 \texttt{T20ID10G0\_00002.ply}.
.4 \texttt{...}.
.4 \texttt{T20ID10G0\_00500.ply}.
.3 \texttt{T20ID10P0/}.
.3 \texttt{T20ID10W0/}.
.3 \texttt{T20ID10R0/}.
.3 \texttt{...}.
.3 \texttt{T40ID60R0/}.
.2 \texttt{pointcloud\_txt/}.
.3 \texttt{T20ID10G0/}.
.4 \texttt{T20ID10G0\_00001.txt}.
.4 \texttt{T20ID10G0\_00002.txt}.
.4 \texttt{...}.
.4 \texttt{T20ID10G0\_00500.txt}.
.3 \texttt{T20ID10P0/}.
.3 \texttt{T20ID10W0/}.
.3 \texttt{T20ID10R0/}.
.3 \texttt{...}.
.3 \texttt{T40ID60R0/}.
}
\end{tcolorbox}
\caption{File organization: 48 design-class subfolders per format, 500 files each.}
\label{fig:mfgnet-file-structure}
\end{figure}

\section{Insights and Notes} 
\label{sec:insights-notes}

Beyond gear design classification and geometric defect detection, MFGNet-Gear supports benchmarking tasks not covered by existing public datasets. Its joint variation in gear design and quality class enables separate evaluation of design classification and defect detection, while also supporting analysis of whether models confuse design-level geometry with defect-level variation. Controlled perturbations in geometric and defect parameters across all 48 design-quality combinations further enable systematic studies of model sensitivity to metrology factors such as point-cloud resolution and noise. The reproducible CAD-to-point-cloud generation pipeline can be extended to additional part designs.

Although the MFGNet-Gear dataset provides a scalable and publicly available benchmark for point-cloud-based defect detection in gears, several limitations remain. First, the dataset is fully synthetic and does not incorporate the sensor noise characteristics of specific 3D metrology instruments. Users adapting models to real scanner data may require domain adaptation before deployment. Second, defect parameter distributions are based on engineering judgment rather than empirical defect inventories from production. Validation against production failure data is needed before real-world deployment. Third, systematic evaluation across the full 12-design $\times$ 4-quality class space remains open. Prior work \cite{mei2024deep} used only subsets, therefore providing only a partial benchmark of the full design space. Fourth, each instance carries a single quality label, whereas gears in service may exhibit concurrent failure modes. Multi-label extensions are left for future work.

\section{Source Code and Scripts} 
\label{sec:source-code}

The SDG pipeline, including SolidWorks macros, design tables, Python sampling scripts, and the validation script, is publicly available through the GitHub repository \url{https://github.com/AliceRSMei/MFGNet-Gear}. The pipeline uses SolidWorks 2021 for CAD generation and mesh export, and relies on Open3D 0.17.0 \cite{zhou2018open3d} and NumPy 1.21.0 \cite{harris2020array} for point-cloud sampling and validation.

\section*{Acknowledgements and Interests}

The authors thank Peter Cerda of the Deep Blue Repository and Research Data Services at the University of Michigan Library for his assistance.

Ruo-Syuan Mei designed and implemented the data generation pipeline, organized the dataset, and wrote the manuscript. Chenhui Shao conceptualized the research, supervised the work, and reviewed the manuscript.

Ruo-Syuan Mei gratefully acknowledges financial support from the Rackham Graduate School at the University of Michigan through the Chia-Lun Lo International Student Fellowship and the Rackham Predoctoral Fellowship.

The authors declare that they have no conflicts of interest.

The authors would like to acknowledge the use of Claude in the preparation of this manuscript. The tool was used for improving wording and correcting English mistakes to improve the flow of the manuscript. The final content was reviewed and approved by the authors.

\bibliographystyle{unsrt}  
\bibliography{references}

\end{document}